\documentclass[11pt]{article}
\usepackage{times}
\usepackage{latexsym}
\usepackage[T1]{fontenc}
\usepackage[utf8]{inputenc}
\usepackage{microtype}
\usepackage{graphicx}
\usepackage{subcaption}
\usepackage[most]{tcolorbox}
\usepackage{enumitem}
\usepackage{array}
\usepackage{adjustbox}
\usepackage{multirow}
\usepackage{hhline}
\usepackage{times}
\usepackage{xspace}
\usepackage{latexsym}
\usepackage{makecell}
\usepackage{soul}
\usepackage[normalem]{ulem}

\usepackage{adjustbox}
\usepackage[T1]{fontenc}
\usepackage[utf8]{inputenc}
\usepackage{microtype}
\usepackage{stfloats}
\usepackage{subcaption}
\usepackage{graphicx}
\usepackage{array}
\usepackage{booktabs}
\usepackage{colortbl}  
\usepackage{subcaption}
\usepackage{ACL2023}

\title{BETA-Labeling for Multilingual Dataset Construction in Low-Resource IR}

\author{\textbf{Md. Najib Hasan}\textsuperscript{1}, \textbf{Mst. Jannatun Ferdous Rain}\textsuperscript{2}, \textbf{Fyad Mohammed}\textsuperscript{3}, \textbf{Nazmul Siddique}\textsuperscript{4}\\
\textsuperscript{1} Wichita State University,
\textsuperscript{2} Begum Rokeya University, Rangpur\\
\textsuperscript{3} Khulna University of Engineering \& Technology, \textsuperscript{4} Ulster University\\
{\tt mxhasan39@shockers.wichita.edu, jannatulferdousrain@gmail.com}\\ 
{\tt fyad4549@gmail.com, nh.siddique@ulster.ac.uk}}

\begin{document}
\maketitle
\begin{abstract}
  IR in low-resource languages remains limited by the scarcity of high-quality, task-specific annotated datasets. Manual annotation is expensive and difficult to scale, while using large language models (LLMs) as automated annotators introduces concerns about label reliability, bias, and evaluation validity. This work presents a Bangla IR dataset constructed using a BETA-labeling framework involving multiple LLM annotators from diverse model families. The framework incorporates contextual alignment, consistency checks, and majority agreement, followed by human evaluation to verify label quality. Beyond dataset creation, we examine whether IR datasets from other low-resource languages can be effectively reused through one-hop machine translation. Using LLM-based translation across multiple language pairs, we experimented on meaning preservation and task validity between source and translated datasets. Our experiment reveal substantial variation across languages, reflecting language-dependent biases and inconsistent semantic preservation that directly affect the reliability of cross-lingual dataset reuse.
Overall, this study highlights both the potential and limitations of LLM-assisted dataset creation for low-resource IR. It provides empirical evidence of the risks associated with cross-lingual dataset reuse and offers practical guidance for constructing more reliable benchmarks and evaluation pipelines in low-resource language settings.
\end{abstract}

\section{Introduction}
IR systems rely on task-specific, accurately annotated datasets to support meaningful evaluation and comparison. For high-resource languages (HRLs), this foundation is well established through carefully curated benchmarks \cite{voorhees2005trec} that enable reproducible experimentation and stable performance assessment \cite{thakur2021beir}. Low-resource languages (LRLs), however, remain largely underserved \cite{ponti2020xcopa, joshi2020state}. The lack of reliable datasets forces researchers to adopt ad hoc evaluation settings or depend on translated resources, creating inconsistencies that weaken comparability \cite{conneau2018xnli} and raise concerns about evaluation validity \cite{kocmi2022toship}.

\begin{figure*}[!t]
\vspace{-3mm}
  \centering
  \begin{minipage}[t]{1\textwidth}
    \begin{subfigure}[t]{\linewidth}
      \includegraphics[width=1\textwidth]{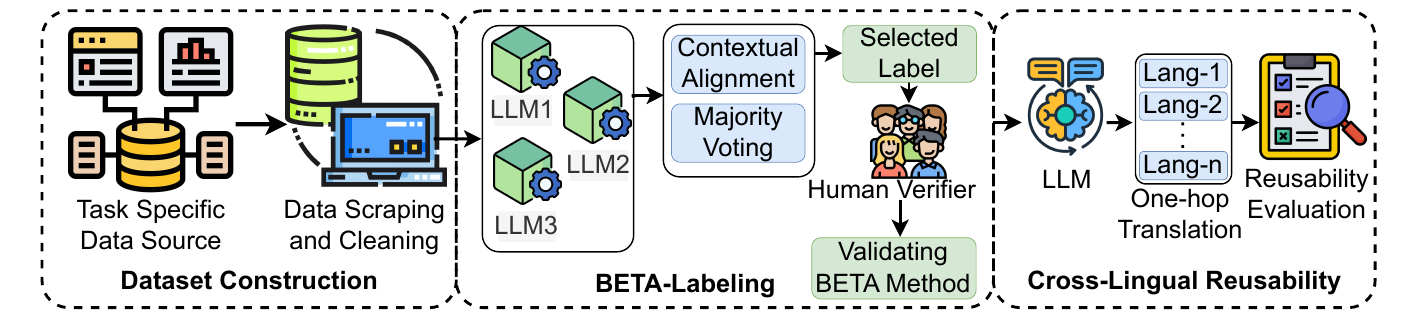}
    \end{subfigure}
  \end{minipage}%
\vspace{-0.5em}
  \caption{Overview of BETA-Labeling and LRL-to-LRL cross-lingual reusability evaluation.}
  \label{fig:safed}
  \vspace{-1.1em}
\end{figure*}

Practical limitations constrain the construction of IR datasets for LRLs. Manual annotation demands linguistic expertise, domain knowledge, and substantial time investment, making large-scale dataset creation difficult to sustain \cite{snow2008cheap}. As a result, recent work has explored LLMs as automated annotators \cite{gilardi2023chatgpt}. While appealing in terms of scalability, the use of LLMs introduces unresolved questions about label reliability in LRL contexts \cite{zhang2023annotators}. Semantic inconsistency \cite{hewitt2023measuring}, implicit meaning distortion \cite{ji2023hallucination, das2025investigating}, and annotator bias \cite{das2024investigating} become particularly problematic when LLM-generated labels are treated as ground truth without systematic validation.

Dataset scarcity has also led to the widespread practice of reusing datasets across languages through machine translation \cite{conneau2018xnli,10.1145/3626772.3657867}. This approach rests on the assumption that translation preserves task semantics, allowing datasets created in one language to remain valid in another. In practice, this assumption is rarely examined, especially for LRL-to-LRL translation \cite{freitag2021bleu,agrawal-etal-2024-translation}. Translation quality varies substantially across language pairs\cite{licht2022consistenthumanevaluationmachine}, and for IR tasks that depend on intent, relevance, or culturally grounded meaning, even minor semantic shifts can alter label validity \cite{kabir2025semanticlabeldriftcrosscultural}. These distortions are often overlooked by surface-level similarity metrics \cite{zhang2020bertscore}, leading to inflated performance estimates and unreliable conclusions\cite{moghe-etal-2023-extrinsic}. Despite the growing reliance on LLMs for both annotation and translation, there is limited empirical understanding of whether LLM-labeled datasets can be safely reused across languages or how language-pair-dependent bias (LPD-bias) affects semantic preservation. Existing evaluation pipelines largely overlook implicit meaning, allowing systematic errors to persist unnoticed \cite{ji2023hallucination}. This gap poses a fundamental challenge for low-resource IR evaluation, where dataset quality and semantic fidelity are as critical as model performance.

To address these issues, we investigate the reliability of LLM-assisted dataset construction and the risks associated with cross-lingual dataset reuse in low-resource settings (see Figure~\ref{fig:safed}). Specifically, we seek to answer the following research questions:

\noindent \textbf{RQ~1.} How reliably can LLMs generate task-specific labels for LRL datasets?

\noindent \textbf{RQ~2.} Can a dataset created in one LRL be effectively reused in other languages via one-hop translation?

\noindent \textbf{RQ~3.} Do LLMs exhibit LPD-bias?



\section{Related Work}
\subsection{Dataset Construction in Low-Resource Language}
The advancement of IR has historically been driven by the availability of standardized benchmarks, most notably those developed under the Text REtrieval Conference (TREC) framework \cite{voorhees2005trec}, as well as large-scale datasets such as MS MARCO \cite{nguyen2016msmarco} and BEIR \cite{thakur2021beir}. These resources have enabled reproducible evaluation and consistent comparison in high-resource languages. In contrast, low-resource languages lack comparable task-specific IR datasets, forcing researchers to rely on small-scale corpora, translated benchmarks, or task-agnostic resources. For some low-resource construction - Dense retrieval models for Amharic language \cite{10.1145/3726302.3730274}, Labadian Crawler for constructing textual corpora from the web \cite{de-jesus-nunes-2024-data}, CWRCzech, a web ranking dataset for Czech \cite{10.1145/3626772.3657851}, ArabicaQA a large-scale open-domain QA dataset \cite{10.1145/3626772.3657889}, Myanmar XNLI \cite{htet2025myanmarxnlibuildingdataset} and CUTE, an open source dataset for Chinese, Uyghur, Tibetan,English \cite{zhuang-sun-2025-cute}.

Prior studies acknowledge that manual dataset construction in low-resource settings is costly, time-intensive, and prone to annotator subjectivity, particularly for tasks requiring interpretation of intent or implicit meaning \cite{ponti2020xcopa}. Although crowdsourcing has been explored as a partial solution, its scalability and reliability remain limited for linguistically and culturally nuanced data. As a result, the scarcity of reliable datasets continues to be a fundamental bottleneck for low-resource IR research.

\subsection{LLMs as Automated Annotators}
Recent work has explored the use of LLMs as automated annotators to address annotation scarcity. Gilardi et al. \cite{gilardi2023chatgpt} show that ChatGPT can match or exceed crowd-worker performance on selected annotation tasks, while subsequent studies examine LLMs as general-purpose annotators across domains \cite{zhang2023llmannotators}. LLMs can be used as active annotator by using them to active learning loop \cite{zhang-etal-2023-llmaaa}, using GPT-3.5 along with few shot and chain-of-thoughts \cite{he-etal-2024-annollm}.These approaches demonstrate the potential of LLMs to reduce annotation costs and accelerate dataset construction. LARA (LLM-Assisted Relevance Assessments) is a suggested method to balance manual annotation and LLM automation \cite{10.1145/3726302.3729916}.

However, existing work largely focuses on high-resource languages and surface-level annotation tasks, often treating LLM outputs as direct substitutes for human labels. Emerging evidence highlights significant limitations, including prompt sensitivity, model-family variation, and cultural bias, particularly when annotations require implicit reasoning or contextual interpretation \cite{ji2023hallucination}. These issues are amplified in low-resource languages, where training data scarcity and linguistic diversity increase semantic instability. Unlike prior studies that rely on single-model annotation, our work adopts a multi-model BETA-labeling approach with explicit agreement mechanisms and human verification.

\subsection{Reuse of Cross-Lingual Dataset}
Cross-lingual IR has long relied on machine translation to extend datasets beyond their original language \cite{conneau2018xnli}. This practice assumes that translation preserves task semantics, enabling dataset reuse without reannotation. While this assumption may hold for closely related or high-resource language pairs, recent work demonstrates that translation quality varies substantially across languages, especially for low-resource and linguistically distant pairs \cite{kocmi2022toship}. Wang et al. proposed a noise-resistant self-distillation method with multiple views in machine translation\cite{10.1145/3503161.3548003}. Manafi et al. investigates a cross-lingual transfer capabilities on 13 pair of languages \cite{manafi-krishnaswamy-2024-cross}.

Standard translation metrics such as BLEU and ROUGE are widely used but have been criticized for their limited sensitivity to semantic adequacy \cite{freitag2021bleu}. Studies show that translations may score highly on lexical similarity while failing to preserve pragmatic or contextual meaning, undermining task validity. Despite these concerns, systematic evaluation of low-resource–to–low-resource dataset reuse remains underexplored, particularly in IR settings where implicit meaning is critical.

\subsection{Implicit Meaning, Hallucination, and Evaluation Gaps}
Implicit meaning plays a central role in IR tasks such as relevance judgment and intent classification. Prior work in natural language generation and translation has documented the phenomenon of hallucination, where models produce fluent but semantically incorrect outputs \cite{ji2023hallucination}. Hewitt et al. \cite{hewitt2023measuring} further show that hallucinations may persist even when semantic similarity metrics remain high, revealing a disconnect between surface-level alignment and true meaning preservation. In IR evaluation, these issues are compounded by the widespread use of metrics that are insensitive to implicit meaning distortion. As a result, translated datasets may yield overly optimistic evaluations while masking systematic semantic errors and LPD-bias. Existing benchmarks rarely account for these risks, leaving a critical gap in low-resource IR evaluation methodology.
\section{Problem Statement}
To examine the reliability of LLM-generated task-specific labels in a low-resource language (LRL) setting, we evaluate a group of LLMs selected from widely used model families. The models are used to predict labels, and final labels are determined through majority voting. 

To assess the effectiveness of LRL datasets for cross-lingual one-hop translation, we apply the same set of LLMs to two datasets, Bangla\_Lite and Bangla\_Culture, and evaluate translation quality using Cosine Similarity, BLEU, and METEOR. The experiments are conducted across five languages: Bangla, English, Chinese, Arabic, and Russian.

To study language-pair-dependent bias, we further evaluate cross-lingual one-hop implicit context transfer using the same LLMs and datasets. Performance is measured using BERTScore, cosine similarity, and Jaccard similarity across the same set of languages: Bangla, English, Chinese, Arabic, and Russian.
\section{Methodology}

\begin{figure*}[!t]
\vspace{-3mm}
  \centering
  \begin{minipage}[t]{0.90\textwidth}
    \begin{subfigure}[t]{\linewidth}
      \includegraphics[width=1\textwidth]{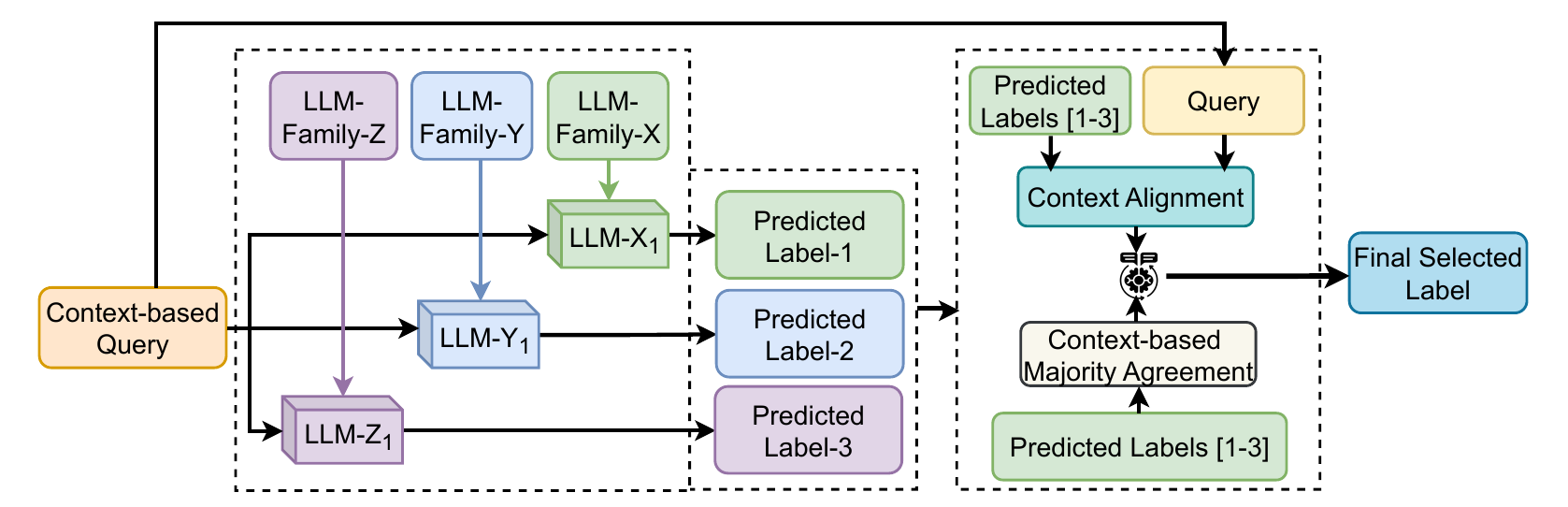}
    \end{subfigure}
  \end{minipage}%
\vspace{-0.5em}
  \caption{BETA-labeling framework to annotate implicit feature-based data in low-resource languages.}
  \label{fig:pipeline}
\end{figure*}

We begin our experiment by collecting task-specific contextual data in a low-resource source language, ``Bangla,'' through documentaries, articles, and literary content from websites and other documents written by native writers. The data are preprocessed to remove noise and normalized to align with the target IR task and label space. Each instance is treated as a query requiring implicit meaning interpretation for label assignment.

\footnotetext{Open-source dataset: \url{https://github.com/MdNajibHasan/Beta_labeling}}

\subsection{Experimental Setup}
\subsubsection{Dataset Details}

\begin{table}[t]
\centering
\setlength{\tabcolsep}{1.5pt}
\begin{adjustbox}{width=\linewidth,center}
\label{tab:dataset_overview}
\begin{tabular}{l|c|c|c|c}
\hline
\textbf{Dataset} & \textbf{\# Samples} & \textbf{Avg. Length} & \textbf{\# Sub-domains} & \textbf{\# Regions} \\
\hline
Bangla\_Lite & 3,822 & 250 tokens & 6 & - \\
Bangla\_Culture & 3,086 & 120 tokens & 4 & 6 \\
\hline
\textbf{Total} & 6,908 & 185 tokens & 10 & 6 \\
\hline
\end{tabular}
\end{adjustbox}
\caption{Overview of the proposed Bangla dataset}
\vspace{-3.1mm}
\end{table}

The design of our dataset is motivated by a core challenge in low-resource IR evaluation: implicit meaning preservation during cross-lingual reuse, particularly when datasets are translated between low-resource languages. Implicit meaning often conveyed through cultural references, figurative language, or contextual knowledge is especially vulnerable to semantic distortion under machine translation. To study this phenomenon in a controlled manner, we construct a task-specific dataset centered on domains where implicit meaning plays a critical role. We focus on two primary domains: (i) Literature (Bangla\_Lite\footnotemark) and (ii) History and Culture (Bangla\_culture\footnotemark). These domains are selected because their interpretation relies heavily on contextual, cultural, and non-literal semantics, making them well-suited for analyzing semantic drift and evaluation reliability in cross-lingual settings.

\textbf{Bangla\_Lite} subset consists of 3,822 samples, covering six major sub-domains: (1) idioms and phrases, (2) music-related text, (3) poems, (4) novels, (5) drama and dialogue, and (6) statements or quotations. Data are collected primarily from publicly available Bangla books and web-based literary resources. These sub-domains are chosen to capture diverse forms of figurative language, narrative context, and stylistic variation.

\textbf{Bangla\_Culture} subset contains 3,086 samples and is designed to represent culturally grounded content across multiple regions and practices. We cover history and cultures of six geographical regions--(1) American, (2) European, (3) Chinese, (4) Indian, (5) African, and (6) Japanese--across four thematic categories: food culture, dress culture, lifestyle, and tradition. Data are sourced from Bangla cultural articles, news reports, and translated books that describe historical and cultural contexts from different regions.

Each data sample is of lengths ranging from 50 to 250 tokens, ensuring sufficient contextual information for interpreting implicit meaning. Importantly, we collect only the source text or query, without explicit annotations of implicit meaning. This design choice reflects realistic dataset construction scenarios and avoids introducing annotation bias at the collection stage. The final dataset size is determined after a rigorous cleaning process, which includes removing erroneous, incomplete, or semantically ambiguous samples. Only high-quality, contextually coherent instances are retained for subsequent labeling and evaluation.

\subsubsection{Models}

\begin{table}[t]
\centering
\setlength{\tabcolsep}{1.5pt}
\begin{adjustbox}{width=\linewidth,center}
\begin{tabular}{l|l|l|l}
\hline
\textbf{Model} & \textbf{Model Family} & \textbf{Model Size} & \textbf{Used For} \\
\hline
GPT-4.1 & OpenAI & Proprietary & BETA-labeling \\
Gemini-3 & Google & Proprietary & BETA-labeling \\
LLaMA-3.3 & Meta LLaMA & 70B & BETA-labeling \\
\hline
GPT-4o mini & OpenAI & Proprietary & Cross-lingual \\
Gemini-2.5 Flash Lite & Google & 8B & Cross-lingual \\
DeepSeek R1 Distilled & DeepSeek & 7B & Cross-lingual \\
\hline
\end{tabular}
\end{adjustbox}
\caption{Overview of models used in different experimental components}
\label{tab:model_overview}
\vspace{-2.1em}
\end{table}

For the BETA-labeling framework, we use three LLMs from different model families—GPT-4.1, Gemini-3, and LLaMA-3.3, for labeling tasks. Each model independently generates labels using the same input context and label space, without access to other model outputs. This design ensures that agreement and disagreement arise from genuine differences in semantic interpretation, allowing BETA-labeling to explicitly analyze annotation uncertainty and model-specific bias.

To evaluate the reliability of cross-lingual dataset generation, we use a separate set of models depending on the dataset. GPT-4o mini, Gemini-2.5 Flash Lite and DeepSeek R1 Distilled (7B) are used for one-hop cross-lingual translation and implicit context transfer on the Bangla\_Lite and Bangla\_culture datasets. These models are used exclusively for cross-lingual experiments and are not involved in BETA-labeling. This enables a clear separation between labeling reliability and cross-lingual transfer evaluation.

\subsection{BETA-labeling}
Automated annotation in low-resource information retrieval often suffers from inconsistent and unreliable labels. To address this issue, we introduce the BETA-labeling framework as shown in Figure~\ref{fig:pipeline}. This structured annotation framework treats model-generated labels as hypotheses to be examined and validated, rather than as ground truth by default. Most existing approaches rely on a single language model or a fixed prompt setup, making the labeling process sensitive to model bias and prompt variation. In contrast, BETA-labeling is designed to explicitly account for annotator diversity, disagreement, and semantic uncertainty. This design is motivated by the observation that, in low-resource IR tasks—especially those involving implicit or contextual meaning—label correctness cannot be determined based solely on fluent output or surface-level similarity.
\subsubsection{Multi-Annotator Design}
Under the BETA-labeling framework, each data instance is annotated independently by three LLMs from different model families: GPT-4.1, Gemini-3, and LLaMA-3.3. These models differ in training data, architecture, and modeling assumptions, which helps reduce the risk of shared or correlated errors. All models receive the same input context and label space, but they generate labels independently, without access to each other’s outputs. This setup encourages diverse semantic interpretations and makes disagreement patterns more informative rather than accidental.
\subsubsection{Contextual Alignment Filtering}
Before labels are combined, each predicted label is evaluated for its contextual alignment with the original query. This step focuses on whether the label captures the underlying intent or meaning of the query, rather than simply overlapping with keywords or topics. Labels that appear fluent but lack a strong semantic connection to the query are given lower priority. This filtering step is particularly important in low-resource settings, where meaning is often expressed indirectly through cultural references, implied intent, or figurative language that cannot be captured through surface matching alone.
\subsubsection{Agreement-Based Selection}
Final label selection is based on majority agreement across the annotating models. However, agreement is treated as an indicator of confidence, not as proof of correctness. When models agree, the label is considered more reliable, but cases of disagreement are retained and analyzed rather than discarded. By preserving disagreement, the framework avoids treating annotation conflicts as noise. Instead, such cases are viewed as signals of ambiguity, semantic complexity, or potential model bias, all of which are important for understanding the limits of automated labeling.

\subsection{Cross-Lingual Dataset Generation}
Rather than constructing separate datasets for each language, we examine whether a dataset labeled in one LRL can be reused across languages through one-hop machine translation. We consider three additional languages commonly underrepresented in IR benchmarks (Chinese, Russian, and Arabic) and one high-resource language (English). Three LLM-based translators—GPT-4o mini, Gemini 2.5 Flash Lite, and DeepSeek R1 Distilled 7B—are used to perform n×n cross-lingual translation (from n languages to all other languages) across all language-pair combinations using a single translation hop; for example, we translate our proposed dataset from Bangla to Chinese using GPT-4o mini. Further, we investigate cross-lingual effectiveness, which is evaluated along two dimensions: (i) translation quality, measured using cosine similarity, BLEU, METEOR; and (ii) implicit meaning preservation, assessed using BERTScore, cosine similarity, and Jaccard similarity by comparing the source and translated instances for task validity retention.
Finally, we analyze LPD bias by examining performance variance across translation directions, highlighting how linguistic asymmetry affects semantic preservation and the reliability of cross-lingual dataset reuse.

\section{Results}

\begin{table*}[htbp]
\centering
\small
\resizebox{\textwidth}{!}{%
\begin{tabular}{l|c|c|c|c|c|c|c|c|c|c|c|c|c|c|c|c|c|c}
\hline
\textbf{\multirow{3}{*}{\makecell[l]{Language\\Pair}}} & \multicolumn{9}{c|}{Bangla\_Lite} & \multicolumn{9}{c}{Bangla\_Culture}\\
\cline{2-19}
& \multicolumn{3}{c|}{GPT-4o Mini} & \multicolumn{3}{c|}{Gemini 2.5 Flash Lite} & \multicolumn{3}{c|}{DeepSeek R1 7B} & \multicolumn{3}{c|}{GPT-4o Mini} & \multicolumn{3}{c|}{Gemini 2.5 Flash Lite} & \multicolumn{3}{c}{DeepSeek R1 7B}\\
\cline{2-19}
& \textbf{CS} & \textbf{BL} & \textbf{ME} & \textbf{CS} & \textbf{BL} & \textbf{ME} & \textbf{CS} & \textbf{BL} & \textbf{ME} & \textbf{CS} & \textbf{BL} & \textbf{ME} & \textbf{CS} & \textbf{BL} & \textbf{ME} & \textbf{CS} & \textbf{BL} & \textbf{ME}\\
\hline
BN $\rightarrow$ EN & 78.9 & 12.3 & 52.4 & 78.0 & 25.0 & 51.2 & 23.4 & 60.0 & 15.5 & 93.1 & 47.4 & 71.8 & 92.6 & 76.3 & 67.5 & 33.3 & 45.9 & 21.4 \\
BN $\rightarrow$ ZH & 72.7 & 3.16 & 1.50 & 69.5 & 10.8 & 1.39 & 30.0 & 7.04 & 1.60 & 93.4 & 64.2 & 6.16 & 91.4 & 39.1 & 5.68 & 35.1 & 0.89 & 1.25   \\
BN $\rightarrow$ AR & 36.4 & 33.0 & 33.1 & 36.4 & 11.5 & 29.3 & 46.9 & 8.89 & 5.27 & 75.6 & 30.7 & 55.0 & 74.8 & 24.7 & 48.6 & 45.8 & 5.15 & 3.94   \\
BN $\rightarrow$ RU & 71.7 & 8.01 & 44.4 & 70.6 & 55.0 & 42.4 & 31.1 & 5.97 & 5.68 & 92.5 & 38.8 & 63.1 & 91.8 & 83.7 & 58.0 & 35.2 & 11.0 & 5.82  \\
\hline
EN $\rightarrow$ BN & 82.8 & 2.47 & 22.7 & 81.5 & 88.8 & 25.4 & 40.5 & 2.58 & 2.98 & 86.1 & 90.4 & 35.8 & 85.0 & 54.2 & 38.4 & 32.0 & 2.52 & 2.93   \\
EN $\rightarrow$ ZH & 73.6 & 1.55 & 1.87 & 70.1 & 77.4 & 2.14 & 44.6 & 0.94 & 0.89 & 93.0 & 46.3 & 5.53 & 91.9 & 82.3 & 5.46 & 54.8 & 0.47 & 1.28   \\
EN $\rightarrow$ AR & 39.3 & 6.66 & 30.5 & 38.9 & 13.3 & 29.6 & 40.6 & 7.59 & 4.36 & 75.0 & 61.0 & 49.1 & 74.7 & 22.0 & 46.3 & 48.2 & 15.3 & 3.89  \\
EN $\rightarrow$ RU & 73.2 & 9.30 & 43.2 & 72.7 & 65.1 & 43.4 & 43.6 & 11.9 & 6.66 & 92.3 & 69.6 & 58.7 & 91.7 & 51.8 & 56.1 & 55.0 & 29.9 & 7.45  \\
\hline
ZH $\rightarrow$ BN & 82.1 & 2.13 & 20.7 & 80.8 & 71.1 & 23.4 & 42.1 & 17.0 & 3.48 & 84.4 & 57.1 & 32.2 & 81.5 & 73.8 & 33.8 & 37.1 & 4.77 & 2.39   \\
ZH $\rightarrow$ EN & 76.1 & 9.77 & 44.3 & 75.2 & 42.8 & 44.4 & 41.3 & 31.4 & 16.7 & 90.2 & 67.6 & 61.7 & 89.9 & 78.0 & 59.4 & 56.9 & 26.8 & 23.9 \\
ZH $\rightarrow$ AR & 36.5 & 5.21 & 25.9 & 36.4 & 10.4 & 25.1 & 42.6 & 2.13 & 4.17 & 73.9 & 49.6 & 44.7 & 73.4 & 98.9 & 41.6 & 51.8 & 23.2 & 3.65  \\
ZH $\rightarrow$ RU & 70.1 & 6.88 & 37.0 & 69.4 & 25.7 & 37.5 & 43.9 & 18.5 & 6.56 & 90.7 & 56.4 & 52.8 & 90.4 & 68.7 & 50.4 & 58.5 & 35.3 & 6.58  \\
\hline
AR $\rightarrow$ BN & 82.9 & 2.32 & 23.2 & 83.1 & 49.8 & 26.6 & 43.1 & 1.92 & 2.31 & 85.2 & 19.0 & 34.2 & 83.9 & 79.1 & 36.5 & 40.7 & 1.74 & 2.24   \\
AR $\rightarrow$ EN & 78.4 & 11.9 & 49.0 & 78.1 & 61.5 & 49.9 & 21.5 & 66.4 & 14.2 & 91.1 & 78.9 & 65.0 & 90.9 & 81.5 & 63.4 & 34.0 & 27.3 & 22.6 \\
AR $\rightarrow$ ZH & 73.6 & 43.7 & 1.72 & 71.7 & 29.9 & 1.85 & 27.4 & 0.82 & 0.89 & 92.5 & 57.5 & 6.38 & 91.1 & 68.7 & 5.90 & 36.4 & 1.02 & 0.94   \\
AR $\rightarrow$ RU & 72.6 & 24.7 & 41.8 & 71.9 & 17.7 & 42.5 & 33.7 & 18.9 & 6.97 & 91.3 & 73.9 & 56.6 & 90.7 & 75.9 & 54.5 & 36.9 & 23.2 & 6.96  \\
\hline
RU $\rightarrow$ BN & 83.4 & 2.56 & 23.8 & 83.6 & 46.9 & 27.1 & 43.1 & 4.31 & 3.25 & 91.8 & 97.7 & 65.8 & 84.9 & 35.9 & 36.9 & 38.4 & 2.75 & 3.18   \\
RU $\rightarrow$ EN & 80.0 & 12.7 & 50.8 & 78.7 & 57.4 & 50.3 & 35.3 & 94.0 & 18.0 & 85.7 & 88.1 & 34.2 & 91.4 & 71.5 & 63.6 & 48.9 & 17.1 & 26.1 \\
RU $\rightarrow$ ZH & 73.3 & 37.4 & 1.61 & 71.8 & 73.2 & 1.83 & 32.4 & 1.27 & 1.02 & 75.0 & 87.1 & 47.9 & 91.9 & 70.2 & 5.69 & 49.4 & 1.30 & 1.32   \\
RU $\rightarrow$ AR & 39.0 & 34.2 & 30.8 & 37.8 & 19.2 & 29.8 & 41.9 & 9.61 & 5.32 & 92.7 & 67.4 & 5.73 & 74.2 & 21.1 & 45.3 & 45.4 & 3.95 & 3.84   \\
\hline
\end{tabular}
}
\caption{Cross-lingual one-hop translation evaluation of three SOTA LLMs (GPT-4o mini, Gemini 2.5 Flash Lite, and DeepSeek R1 7B) on the Bangla\_Lite and Bangla\_History datasets. In this table, CS, BL, and ME denote Cosine Similarity, BLEU, and METEOR, respectively. BN, EN, ZH, AR, and RU represent Bangla, English, Chinese, Arabic, and Russian, respectively.}
\label{tab:cross_lingual_metrics}
\end{table*}


In Table \ref{tab:cross_lingual_metrics}, we evaluate explicit one-hop cross-lingual translation quality on Bangla\_Lite and Bangla\_Culture. On Bangla\_Lite, GPT-4o Mini, and Gemini 2.5 Flash Lite achieve cosine similarity between 70–83\% for Bangla-to-English, Bangla-to-Chinese, and Bangla-to-Russian. This result indicates strong surface-level semantic alignment. However, for form-sensitive metrics, the scores aren't consistent for the same translations: BLEU remains consistently low (mostly below 15\%), and METEOR rarely exceeds 55\%, revealing weak lexical and structural correspondence. On Bangla\_Culture, cosine similarity increases much higher. Scores exceed 90\% for English-centric and Russian-centric directions, and METEOR is frequently above 60\%. Despite this improvement, BLEU continues to vary widely across language pairs. Variations across BLEU and METEOR scores suggest that the translation behavior is rather unstable. DeepSeek R1 7B performs markedly worse across both subsets, with cosine similarity often below 45\% on Bangla\_Lite and METEOR commonly below 25\%. These results show that even when LLMs generate semantically fluent outputs, explicit translation quality alone does not guarantee reliable task-specific labels in LRLs \textbf{\underline{(Answers RQ1)}}.

Table \ref{tab:implicit} highlights a pronounced degradation when implicit meaning is evaluated. On Bangla\_Lite, GPT-4o Mini and Gemini 2.5 Flash Lite produce BERTScores in a narrow range of 66–69\%, yet Jaccard similarity lowers to near 0\% for most non-English targets. Cosine similarity remains moderate (58–64\%), indicating that even though the semantics are shared, the contextual structure or implicit meaning isn't guaranteed to be preserved. A similar pattern holds for Bangla\_Culture. BERTScore increases slightly (70–73\%) across language pairs. DeepSeek R1 7B shows the weakest overall performance, with BERTScores below 58\% and negligible overlap across all implicit metrics. These findings demonstrate that one-hop translation is fundamentally insufficient for preserving implicit meaning, implying that direct reuse of the cross-lingual dataset is unreliable where low-resource IR tasks depend on the context\textbf{\underline{(Answers RQ2)}}.

\begin{table*}[htbp]
\centering
\small

\resizebox{\textwidth}{!}{%
\begin{tabular}{l|c|c|c|c|c|c|c|c|c|c|c|c|c|c|c|c|c|c}
\hline
\textbf{\multirow{3}{*}{\makecell[l]{Language\\Pair}}} & \multicolumn{9}{c|}{Bangla\_Lite} & \multicolumn{9}{c}{Bangla\_Culture}\\
\cline{2-19}
& \multicolumn{3}{c|}{GPT-4o Mini} & \multicolumn{3}{c|}{Gemini 2.5 Flash Lite} & \multicolumn{3}{c|}{DeepSeek R1 7B} & \multicolumn{3}{c|}{GPT-4o Mini} & \multicolumn{3}{c|}{Gemini 2.5 Flash Lite} & \multicolumn{3}{c}{DeepSeek R1 7B}\\
\cline{2-19}
& \textbf{BS} & \textbf{CS} & \textbf{JS} & \textbf{BS} & \textbf{CS} & \textbf{JS} & \textbf{BS} & \textbf{CS} & \textbf{JS} & \textbf{BS} & \textbf{CS} & \textbf{JS} & \textbf{BS} & \textbf{CS} & \textbf{JS} & \textbf{BS} & \textbf{CS} & \textbf{JS}\\
\hline
BN $\rightarrow$ EN & 67.6 & 59.5 & 10.9 & 66.9 & 57.3 & 9.7 & 56.1 & 38.3 & 6.12 & 72.3 & 71.0 & 13.7 & 68.5 & 66.4 & 11.4 & 57.6 & 44.6 & 7.92 \\
BN $\rightarrow$ ZH & 68.4 & 63.3 & 0.01 & 67.5 & 60.1 & 0.0 & 54.9 & 34.1 & 0.16 & 73.2 & 77.2 & 0.02 & 69.5 & 72.5 & 0.02 & 56.5 & 34.9 & 0.09   \\
BN $\rightarrow$ AR & 67.5 & 43.2 & 6.40 & 67.5 & 41.8 & 4.8 & 53.3 & 41.1 & 0.56 & 70.8 & 66.1 & 8.64 & 68.8 & 61.6 & 6.33 & 56.9 & 39.8 & 0.68   \\
BN $\rightarrow$ RU & 67.1 & 60.3 & 7.45 & 67.2 & 59.6 & 6.1 & 52.6 & 34.1 & 0.72 & 71.0 & 76.3 & 8.71 & 68.5 & 71.2 & 6.21 & 57.3 & 36.4 & 0.42  \\
\hline
EN $\rightarrow$ BN & 69.5 & 62.4 & 9.39 & 69.6 & 64.1 & 7.9 & 56.7 & 38.7 & 1.17 & 73.3 & 63.1 & 11.9 & 71.3 & 64.2 & 9.16 & 57.6 & 40.6 & 1.17   \\
EN $\rightarrow$ ZH & 68.7 & 64.6 & 0.03 & 67.4 & 59.9 & 0.0 & 56.2 & 34.7 & 0.12 & 73.2 & 77.4 & 0.01 & 69.4 & 72.8 & 0.02 & 57.4 & 38.6 & 0.05   \\
EN $\rightarrow$ AR & 67.8 & 45.0 & 6.75 & 67.5 & 42.8 & 5.2 & 50.2 & 38.6 & 0.55 & 70.7 & 65.8 & 8.51 & 68.8 & 61.7 & 6.29 & 56.8 & 42.2 & 0.64  \\
EN $\rightarrow$ RU & 67.2 & 62.1 & 7.74 & 67.3 & 60.6 & 6.3 & 52.8 & 38.3 & 1.16 & 71.0 & 76.3 & 8.86 & 68.6 & 71.5 & 6.31 & 54.5 & 44.7 & 1.00  \\
\hline
ZH $\rightarrow$ BN & 69.1 & 63.1 & 9.03 & 69.5 & 65.2 & 7.8 & 57.7 & 38.2 & 1.39 & 73.0 & 62.6 & 11.5 & 70.9 & 63.6 & 8.78 & 56.7 & 42.5 & 1.38   \\
ZH $\rightarrow$ EN & 67.4 & 58.9 & 10.5 & 66.5 & 56.0 & 9.1 & 56.4 & 42.1 & 6.16 & 71.9 & 70.3 & 13.3 & 68.4 & 65.8 & 11.2 & 57.6 & 49.3 & 7.77 \\
ZH $\rightarrow$ AR & 67.4 & 44.3 & 6.25 & 67.4 & 42.2 & 4.8 & 52.7 & 41.1 & 0.73 & 70.4 & 65.7 & 8.31 & 68.7 & 61.2 & 6.20 & 54.5 & 41.4 & 0.74  \\
ZH $\rightarrow$ RU & 66.9 & 60.3 & 7.42 & 67.0 & 59.1 & 6.1 & 53.0 & 40.3 & 1.59 & 70.6 & 75.8 & 8.54 & 68.3 & 71.2 & 6.25 & 54.5 & 47.1 & 1.22  \\
\hline
AR $\rightarrow$ BN & 69.4 & 62.0 & 9.29 & 69.8 & 65.2 & 8.3 & 55.9 & 33.3 & 0.90 & 73.1 & 62.9 & 11.6 & 71.0 & 63.8 & 8.97 & 56.6 & 38.9 & 0.87   \\
AR $\rightarrow$ EN & 67.6 & 59.9 & 10.9 & 66.6 & 57.6 & 9.8 & 54.8 & 32.9 & 5.37 & 72.0 & 70.6 & 13.4 & 68.4 & 66.0 & 11.3 & 56.5 & 40.6 & 6.97 \\
AR $\rightarrow$ ZH & 68.7 & 64.3 & 0.03 & 67.5 & 61.3 & 0.0 & 55.2 & 27.7 & 0.12 & 73.0 & 77.1 & 0.00 & 69.3 & 71.8 & 0.00 & 57.2 & 32.8 & 0.09   \\
AR $\rightarrow$ RU & 67.3 & 61.9 & 7.80 & 67.1 & 60.1 & 6.3 & 51.9 & 32.3 & 0.89 & 70.9 & 75.9 & 8.73 & 68.4 & 70.5 & 6.15 & 59.2 & 35.2 & 0.82  \\
\hline
RU $\rightarrow$ BN & 69.4 & 63.1 & 9.39 & 69.9 & 66.0 & 8.3 & 56.1 & 33.3 & 1.00 & 72.1 & 70.8 & 13.5 & 71.1 & 63.5 & 9.15 & 56.9 & 40.6 & 0.90   \\
RU $\rightarrow$ EN & 67.7 & 60.3 & 10.9 & 66.7 & 58.0 & 9.8 & 55.3 & 39.9 & 5.86 & 73.2 & 63.0 & 11.6 & 68.5 & 66.5 & 11.3 & 56.8 & 45.2 & 7.58 \\
RU $\rightarrow$ ZH & 68.6 & 64.3 & 0.01 & 67.6 & 61.5 & 0.0 & 54.3 & 31.7 & 0.10 & 70.7 & 65.8 & 8.50 & 69.4 & 72.9 & 0.02 & 57.3 & 36.4 & 0.07   \\
RU $\rightarrow$ AR & 67.7 & 44.3 & 6.69 & 67.6 & 42.6 & 5.1 & 57.7 & 40.4 & 0.68 & 73.0 & 77.3 & 0.01 & 68.8 & 61.3 & 6.34 & 58.3 & 41.4 & 0.62   \\
\hline
\end{tabular}
}
\caption{Cross-lingual one-hop implicit context evaluation of three SOTA LLMs (GPT-4o mini, Gemini 2.5 Flash Lite, and DeepSeek R1 7B) on the Bangla\_Lite and Bangla\_History datasets. In this table, BS, CS, and JS denote BERTScore, Cosine Similarity, and Jaccard Similarity, respectively. BN, EN, ZH, AR, and RU represent Bangla, English, Chinese, Arabic, and Russian, respectively.}
\label{tab:implicit}
\end{table*}

Across both tables, systematic asymmetries arise depending on the language pair and translation direction. English-to-Bangla and Russian-to-Bangla consistently outperform translations in the reverse direction, and the cosine similarity and METEOR are often 10–15 \% higher. In contrast, Bangla-to-Arabic and Bangla-to-Chinese are shown to have drastic performance drops, proven by the frequency of cosine similarity below 40\%. These gaps persist across GPT-4o Mini, Gemini 2.5 Flash Lite, and DeepSeek R1 7B, indicating that the bias is not model-specific. Evaluation for implicit-context amplifies the difference, which shows implicit intent is unevenly preserved across directions and families, even though the explicit translation is apparently adequate. Together, these results confirm the presence of LPD bias, undermining the assumption that LLM-generated annotations are stable or transferable across LRLs. \textbf{\underline{(Answers RQ3)}}.

\section{Discussion}
This work shows that dataset quality in low-resource IR depends more on meaning than on surface fluency. While LLMs can generate labels and translations at scale, semantic stability is not guaranteed, particularly for tasks that rely on implicit intent and contextual understanding. The BETA-labeling framework improves reliability by treating model outputs as hypotheses rather than ground truth. The disagreements and uncertainty from the multiple families at the surface can be accepted instead of being kept hidden. In single-model labeling, the disagreements are overlooked. Labeling errors are reduced by contextual alignment and majority agreement, but human verification confirms that agreement alone does not ensure correctness. In many cases, culturally figurative or grounded contents show genuine semantic ambiguity rather than noise. Cross-lingual reuse through one-hop translation is proven unreliable in low-resource language settings. The explicit metrics show that performance for some language pairs is acceptable. But the implicit meaning metrics show severe degradation. Frequent performance drops to near zero in case  Jaccard similarity shows that the core intent is often lost while the surface-level meaning appears to be preserved. Clear LPD bias is observed across all models and datasets. English-to-Bangla and Russian-to-Bangla have a higher performance rate than their translation in reverse directions. It shows the structural asymmetries rather than model-specific failures. When Bangla is used as the source language, semantic loss increases sharply, reflecting the limitations of source-side interpretation. For Bangla-to-Arabic and Bangla-to-Chinese translations, we observe the weakest performance. This semantic drift is amplified due to cultural distance, syntactic differences, and limited shared context. The implicit meaning suffers the most in these cases. As a result, fluent translation does not imply the validity of the task. All the findings question the reusability of the translated datasets in IR evaluation. The semantic drift is often masked by surface-level similarity and leads to inflated conclusions. Without explicit checks for implicit meaning, cross-lingual evaluation remains unreliable. Finally, LRL IR datasets are not portable by default; language direction and cultural grounding matter. BETA-labeling provides a safer framework, but human validation remains essential.
\section{Conclusion}
In this study, we examine how reliable LLM-assisted dataset construction is and what the risks of cross-lingual reuse are in low-resource IR. LLMs offer a scalable alternative to manual annotation. However, the experiment confirms that scalability alone is not enough to ensure reliability. The labels and translations may appear fluent. But task-relevant meaning, especially the implicit meaning, can not be captured properly. The strength of the BETA-labeling framework relies on treating the model outputs as hypotheses rather than the ground truth. This framework also exposes disagreement and semantic certainty that would remain hidden. Human verification further confirms that automated agreement is an indicator of confidence, not correctness. Our cross-lingual experiments reveal that one-hop translation is insufficient for reliable dataset reuse in LRLs. Explicit similarity metrics often suggest acceptable transfer, but implicit-context evaluation shows substantial semantic degradation across many language pairs. These failures are not isolated to a single model but persist across architectures and model sizes. We also identify a strong LPD bias that consistently favors HRL-to-LRL directions over the reverse. When LRLs serve as source languages, semantic loss increases sharply, especially for linguistically distant pairs. This undermines the assumption that LLM-generated annotations are stable or transferable across languages. This work highlights a critical evaluation gap in low-resource IR. Relying on surface-level metrics and untranslated benchmarks will not lead to a proper conclusion. In order to evaluate reliably, attention must be given to the implicit meaning, language direction, and cultural context. BETA-labeling offers a more robust path for dataset construction. Cross-lingual reuse can not be guaranteed as a safe method, and it needs to be validated carefully rather than direct adoption driven by convenience.
\section{Limitation}
This work deliberately focuses on a controlled experimental setting, which introduces several limitations. First, the dataset is constructed in Bangla and evaluated through translation into a limited set of target languages. While these languages are chosen to represent diverse scripts, linguistic structures, and cultural distances, the results should not be interpreted as universally representative of all LRLs or IR tasks. Although BETA-labeling reduces reliance on a single annotator, it remains an LLM-assisted framework. Human verification is used to validate label quality, but full human annotation is not feasible at this scale. Consequently, the study does not claim absolute label correctness; instead, it evaluates relative reliability and stability under realistic dataset construction constraints. Implicit meaning preservation is evaluated using automated similarity measures. These metrics are intentionally selected to expose failures that surface-level metrics overlook, but they cannot fully capture pragmatic intent or cultural interpretation. A fully human-driven evaluation of implicit meaning would provide deeper insight, but it lies outside the practical scope of large-scale IR benchmarking. The cross-lingual analysis is restricted to one-hop translation. This choice reflects common practice in dataset reuse, but we will consider alternative strategies, such as pivot-based or multi-hop translation, to mitigate some semantic loss. The goal here is not to exhaust all translation methods but to test the validity of a widely assumed reuse pipeline. Finally, the experiments rely on a fixed set of contemporary LLMs and prompts. While multiple model families are used to reduce correlated bias, the findings should be interpreted as a snapshot of current LLM behavior. As models evolve, the magnitude of observed biases may change, though the underlying risks identified in this study are likely to persist.

\bibliographystyle{acl_natbib}
\bibliography{custom}


\end{document}